%% file: main.tex
\documentclass[letterpaper, 10 pt, conference]{ieeeconf}
\IEEEoverridecommandlockouts
\usepackage{caption}
\usepackage{subcaption}
\usepackage{cite}
\usepackage{amsmath,amssymb,amsfonts}
\usepackage{algorithmic}
\usepackage{graphicx}
\usepackage{textcomp}
\usepackage{xcolor}
\def\BibTeX{{\rm B\kern-.05em{\sc i\kern-.025em b}\kern-.08em
    T\kern-.1667em\lower.7ex\hbox{E}\kern-.125emX}}
\begin{document}

\title{\LARGE \bf Automatic Sign Reading and Localization for Semantic Mapping with an Office Robot}

\author{David Balaban$^{1}$ and Justin Hart$^{2}$
\thanks{$^{1}$David Balaban is with the department of Computer Science, University of Texas at Austin, Austin, Texas USA {\tt\small dbalaban@cs.utexas.edu}}
\thanks{$^{2}$ Justin Hart is with faculty of Computer Science, University of Texas at Austin, Austin, Texas USA {\tt\small hart@cs.utexas.edu}}
}
\maketitle

\begin{abstract}
Semantic mapping is the task of providing a robot with a map of its environment beyond the open, navigable space of traditional Simultaneous Localization and Mapping (SLAM) algorithms by attaching semantics to locations. The system presented in this work reads door placards to annotate the locations of offices. Whereas prior work on this system developed hand-crafted detectors, this system leverages YOLOv5 for sign detection and EAST for text recognition. Placards are localized by computing their pose from a point cloud in a RGB-D camera frame localized by a modified ORB-SLAM. Semantic mapping is accomplished in a post-processing step after robot exploration from video recording. System performance is reported in terms of the number of placards identified, the accuracy of their placement onto a SLAM map, the accuracy of the map built, and the correctness transcribed placard text.
\end{abstract}

\section{Introduction}
Simultaneous Localization And Mapping (SLAM) techniques provide navigational maps that enable robots to autonomously explore an environment. The emerging family of Semantic Mapping techniques add semantic information about locations and objects in the environment to this basic map data. Semantic mapping enables robots to navigate based on a location's semantics rather than simple geometric poses, and by extension allows the robot to reason about locations using automated planning or based on semantic parses of natural language commands~\cite{thomason2020jointly}. Semantic maps can be particularly useful for consumer-facing robotic applications, such as aiding visitors in navigating an unfamiliar building~\cite{Yedidsion_multi_robot} or object retrieval~\cite{Jiang_Walker_Hart_Stone_2019}. Manually labeling map semantics can be tedious or impractical for large maps. This work focuses on the autonomous semantic labeling of a mapped environment.

Buildings often feature signage that annotates important semantics for the benefit of human occupants, such as placards indicating office numbers. Modern Optical Character Recognition (OCR) methods can be used to extract semantic meaning from text found in optical frames. Prior work used hand crafted detectors to identify door placards, register their position on a map, and transcribe their labels~\cite{Hart}~\cite{Case}. These methods require either the construction of a map prior to scanning for placards, or for a human to guide the robot while scanning for placards. The contribution of this work is to construct the map and semantics in post-processing of video RGB-D data recorded during exploration.

This work utilizes RGB-D sensor data collected from a robot equipped with an Azure Kinect camera~\cite{tolgyessy2021evaluation}. Video data was recorded during human-guided exploration of an office floor. From the video recording, a navigation map is built using a version of ORB-SLAM~\cite{mur2017orb} which has been modified to improve performance in the presence of very challenging input, in the form of flat-colored walls with few obvious features to match during visual reconstruction (an important step in the ORB-SLAM pipeline). After the navigation map is built, extracted keyframes are utilized to build a full 3D reconstruction of the hallways from sensor data. At the same time, a trained YOLOv5~\cite{glenn_jocher_2022_7002879} object-detector is used to identify door placards in each frame. The detector used in this study is specifically trained to detect door number placards. If a placard is detected, it is localized in frame and added to the reconstruction. EAST~\cite{zhou2017east} and Tesseract-OCR~\cite{smith2007overview} are used to read and record each placard's text. The text contains information such as room numbers, which then is used to mark up the semantics of room locations on the constructed map.

This paper demonstrates an effective semantic mapping system for recorded RGB-D data of an office hallway which reads and localizes room numbers (Sec.~\ref{sec:mappingSystem}). The success of this work is measured in terms of the number of placards identified, the accuracy of the localization, the ability to make accurate maps in the presence of featureless walls, and the number of placards successfully read.

%%JUSTIN - Up to here is great. Don't change anything if you can avoid it.

\section{Related Work}
Simultaneous Localization and Mapping (SLAM) refers to the problem of exploring an environment, mapping the features of that environment, and placing the robot on the map~\cite{cadena2016past}. Semantic Mapping is the problem of placing semantic labels on map features~\cite{sunderhauf2017meaningful} for efficient task completion or natural communication with humans~\cite{aydemir2011plan}\cite{pronobis2012large}. Semantic SLAM tackles both problems at once, often using semantic labels in the localization process~\cite{bowman2017probabilistic}.

Much of the work on Semantic Mapping and Semantic SLAM, has focused on identifying object labels with the sensor being considered, often with RGB or RGB-D cameras~\cite{nakajima2018efficient}\cite{sunderhauf2017meaningful}\cite{8794344}, although identifying rooms and spaces is also a common problem~\cite{7487234}. Objects are placed onto a map using a representative model ranging from detailed 3D models~\cite{hosseinzadeh2019real}, to bounding ellipsoids and cubes~\cite{8794344}, or a labelled point cloud and voxel grid~\cite{yu2018variational}.

Labels are generally categorized a-priori by type, without discrimination between instances of the same type~\cite{sualeh2019simultaneous}. An alternative is to discover labels through text in the environment that uniquely identify objects or spaces, thereby improving the specificity of the semantic map. Prior work identified text as a useful indicator of planar objects rich in distinct visual features useful for optical SLAM methods~\cite{li2020textslam}\cite{ma2021homography}. Transcribing text has been used for localization on a given map~\cite{wang2015lost}\cite{sadeghi2017ocrapose}. Text recognition as a part of a SLAM system has also been used in self-driving cars and retail spaces~\cite{kenye2020re}\cite{swaminathan2019autonomous}\cite{dworakowski2021robot}. A major difference between street signs and office signs is that the former is designed to be easily read from forward facing fast moving vehicles, while the latter by slow moving pedestrians capable of reorienting their view at will. This difference causes many of the challenges discussed in this paper.

There are two major investigations which we expand upon. The earliest work is Autonomous Sign Reading (ASR) for Semantic Mapping~\cite{Case}, and the other is Pose Registration for Integrated Semantic Mapping (PRISM)~\cite{Hart}. While prior work focuses on design of placard detectors, this paper focuses on efficient discovery.

%JUSTIN - Related work is okay. I made some small removals that you may want to review.

\section{Semantic Mapping System}
\label{sec:mappingSystem}
This semantic mapping system constructs maps from video and depth data (RGB-D video) recorded using a Microsoft Azure Kinect \cite{tolgyessy2021evaluation} mounted on a mobile robot (a University of Texas at Austin Building-Wide Intelligence BWIBot \cite{IJRR17-khandelwal}). 

Recordings are processed in two phases. In the first phase, a SLAM map is constructed with a modified ORB-SLAM~\cite{mur2017orb}. The second phase extracts semantics from each keyframe. The second phrase comprises several steps:

\begin{enumerate}
    \item Detect placards (Regions of Interest - ROIs)
    \item Extract planar features
    \item Identify text lines
    \item Transcribe text to string
    \item Validate string
    \item Map Reconstruction and Projection
    \item Data aggregation
\end{enumerate}

As each placard is observed, the pose and label is recorded. If the label could not be validated, it is left blank. 

\subsection{A Modified ORB-SLAM}
\label{sec:modify}
ORB-SLAM works by identifying distinct visual features known as ORB features~\cite{rublee2011orb} from visual RGB images, and matching these features across a number of video frames. Once these features are matched, it searches for the transformation between frames which best explains the visual position of the matches. ORB-SLAM also performs keyframing for efficient memory usage for storing ORB features. In this way, ORB-SLAM tracks the most current frame in a visual feed with respect to a map's origin frame as long as enough ORB features are matched to previous keyframes. If this tracking is lost, ORB-SLAM creates a new map and begins tracking again. 

Motion blur or frames which lack matching ORB features can cause ORB-SLAM to lose track of the current frame's pose. It is possible for ORB-SLAM to merge maps only when the camera returns to a formerly observed viewpoint.

When the camera returns to a previous viewpoint, ORB-SLAM will either do a loop closure, or a map merge. If the matching frames are in the same map, then loop closure will frequently fix localization errors. However, if ORB-SLAM detects a matching frame on a different map than the one currently being tracked, a map merge is performed, and this opportunity for refined localization is lost.

Mono-colored walls present a danger that ORB-SLAM will be unable to reliably determine the camera-pose that a frame of video was shot from due to a lack of visual features, thus creating flaws in the underlying map. Every time the robot rotates --- such as to face a placard head-on or recover from navigating into a tight space --- there is a good chance ORB-SLAM will lose track. In the Anna Hiss Gymnasium robotics facility --- where there are many beautiful, but unfortunately plain white walls --- a method of merging these maps that supplements the capabilities of standard ORB-SLAM is necessary. It is also best to recover the robot's accurate pose quickly after tracking is lost to avoid missing an opportunity for loop-closure when ORB-SLAM fails to identify sufficient matching features.

To address this, we enhanced ORB-SLAM by matching the full 3D point clouds of the last successfully localized frame in the old map with the first frame (the origin frame) of the new map. An iterative closest point (ICP) algorithm is used to identify the transform which best matches the point clouds~\cite{besl1992method}. By identifying the transform between frames based on point clouds, the new map is localized in the reference frame of the old, and the maps can be merged using ORB-SLAM's out-of-the-box functionality.

This method is computationally expensive, so processing these supplemental map merges needs to be done offline.

\subsection{Detecting placards}
\begin{figure*}
    \centering
    \includegraphics[width=\linewidth]{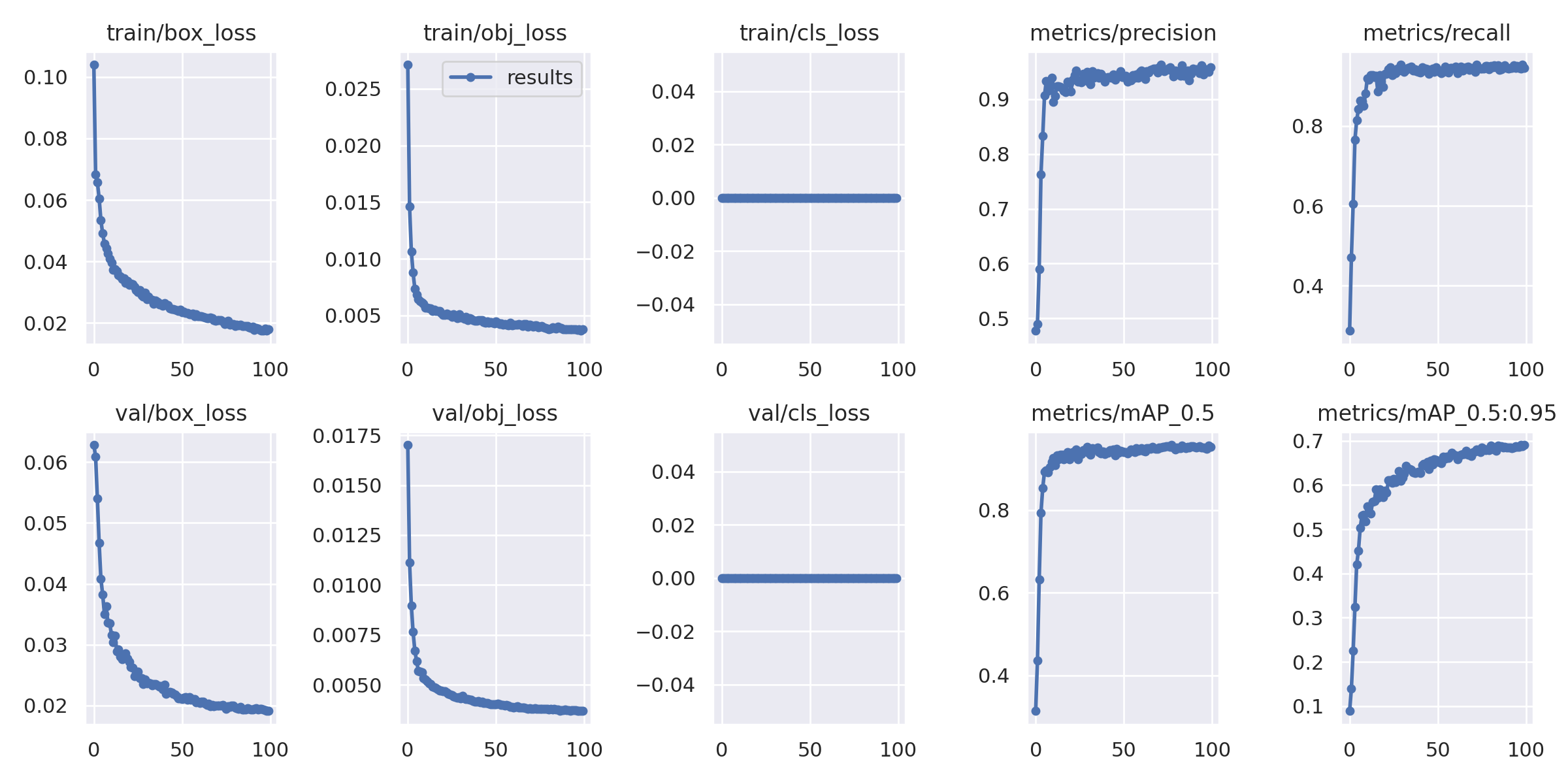}
    \includegraphics[width=\linewidth]{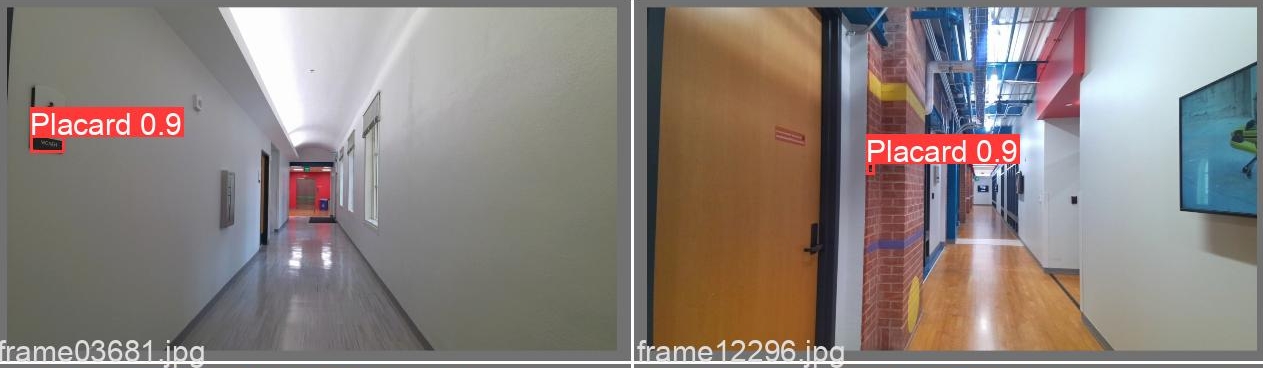}
    \caption{Training metrics by epoch (top) and validation set examples with confidence (bottom)}
    \label{fig:yoloTraining}
\end{figure*}

The You Only Look Once (YOLO) neural network~\cite{Yolo} is utilized to identify placards in frame and extract ROIs in the RGB frame of each ORB-SLAM keyframe. A YOLOv5~\cite{glenn_jocher_2022_7002879} network was re-trained on pictures of placards coming from the right half of the building in Figure~\ref{fig:hand_crafted}, the other half of the building is held out as test set and unseen in the training data. Because the placards are standardized, the system generalizes well to the new space. Figure~\ref{fig:yoloTraining} shows the training metrics over $100$ epochs, along with two example validation images. A confidence threshold of $0.9$ is used to filter false detections.

\begin{figure}
    \centering
    \subfloat[\centering Detected ROI from YOLOv5.]{{\includegraphics[width=.55\linewidth]{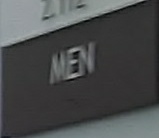} }}%
    \quad
    \subfloat[\centering EASTnet lines, warped ROI.\label{fig:textExtraction:a}]{{\includegraphics[width=.3\linewidth]{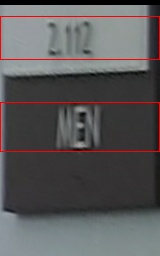} }}%
    \quad
    \subfloat[\centering Binary image, threshold=$85$.\label{fig:textExtraction:b}]{{\includegraphics[width=\linewidth]{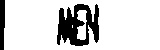} }}%
    \caption{Text extraction steps from YOLOv5 detection.}%
    \label{fig:textExtraction}%
\end{figure}

Figure~\ref{fig:textExtraction} a shows an image patch of a detected placard, the patch edges are defined by the bounding box returned by the trained YOLOv5.

\subsection{Extract Planar Features}
The Azure Kinect uses a time-of-flight camera paired with a 4K RGB camera. These two views are registered to each other through camera calibration, allowing the extraction of a point cloud from the depth camera which is associated with a placard captured in the RGB camera. The point cloud is used to extract the plane in which the placard lies. Given the camera frame's pose from ORB-SLAM, the placard's pose in camera frame is transformed into ORB-SLAM's global frame, thereby localizing the placard.

\subsection{Identify Text Lines}
Given the pose of the placard in the camera frame, the perspective can be warped in the ROI to approximate a head-on view, as shown in Figure~\ref{fig:textExtraction}a. This warping is done to enhance text readability. An Efficient and Accurate Scene Text (EAST)~\cite{zhou2017east} pre-trained detector is used to identify lines of text in each ROI. Tesseract-OCR will ignore colors which differ from the first recognized line in an image. Using EAST, the system subdivides the ROI containing the placard into regions of text with the same color. An example of these subdivisions is shown in Figure~\ref{fig:textExtraction}b, the lines of text are outlined in red.

\subsection{Transcribe Text to String}
Once the image is subdivided by EAST into lines, each line is then passed to Tesseract-OCR~\cite{smith2007overview} to transcribe the text to a character string. The system uses Tesseract's pre-trained English dictionary.

Tesseract works best on binary images, so the images are converted to black and white and a threshold applied before transcription. Several thresholds are taken to generate multiple proposed strings. Threshold values are taken in increments of five on the range $[5,250]$ for an eight-bit color value. Figure~\ref{fig:textExtraction}c shows an example of this thresholding with a value of 85, this value was hand-picked as the most readable.

\subsection{Validate String}
The placards follow a standardized format for indicating room numbers: the single-digit floor number, a decimal point, and a three digit room number. There are a handful of placards where the room name is given in place of the number, these are the restrooms (MEN, WOMEN, GENDER INCLUSIVE) and the stairs (STAIR1, STAIR2). The system validates that the returned string adheres to these standards with a regular expression. The decimal point is optional because it is often hard to identify due to its size. The stair ID is also optional because the distinction is not of much practical use. If successfully validated, the string is recorded as the placard's label.

\subsection{Map Reconstruction and Projection}
The depth-map and camera pose of each keyframe are passed into octomap to perform a full 3D reconstruction of the scene~\cite{hornung13auro}. The system uses a map resolution of $0.03$ m. For ease of visualization the system also projects the reconstruction into a 2D map with vertical cuts applied for the floors and ceiling. Vertical camera drift in ORB-SLAM's localization is corrected by adjusting the frame's vertical coordinate to a constant value.

\subsection{Placard Aggregation}
Once all keyframes have been processed, placards which are not localized to lie on walls are discarded, and those observations which lie within one placard radius ($0.151$m) of each other are grouped together. Among each observation group, the average pose is taken, and the most common identified label adopted. This consistently localizes the placards and semantically labels them onto a navigable SLAM map.

\section{Experimental Design}
\label{sec:recordings}
A video was recorded on an Azure Kinect sensor~\cite{tolgyessy2021evaluation}. The color camera was set to a resolution of $3840\times2160$px @ $15$ fps, and the depth camera was set to a binned wide field of view (WFOV) with $512\times512$px @ $15$ fps. The color camera resolution is set such that placards viewed at far distances and poor angles have the best possible chance of being read. The depth camera is set so WFOV has the best possible overlap with the color camera and the maximum possible depth range for WFOV ($0.25-2.88$m).

\subsection{Recording Features}
In the recording, the robot is driven near continuously making no attempt to view the placards head-on, with two full rotation in-place around the robot's center.

\begin{figure}
    \centering
    \includegraphics[width=\linewidth]{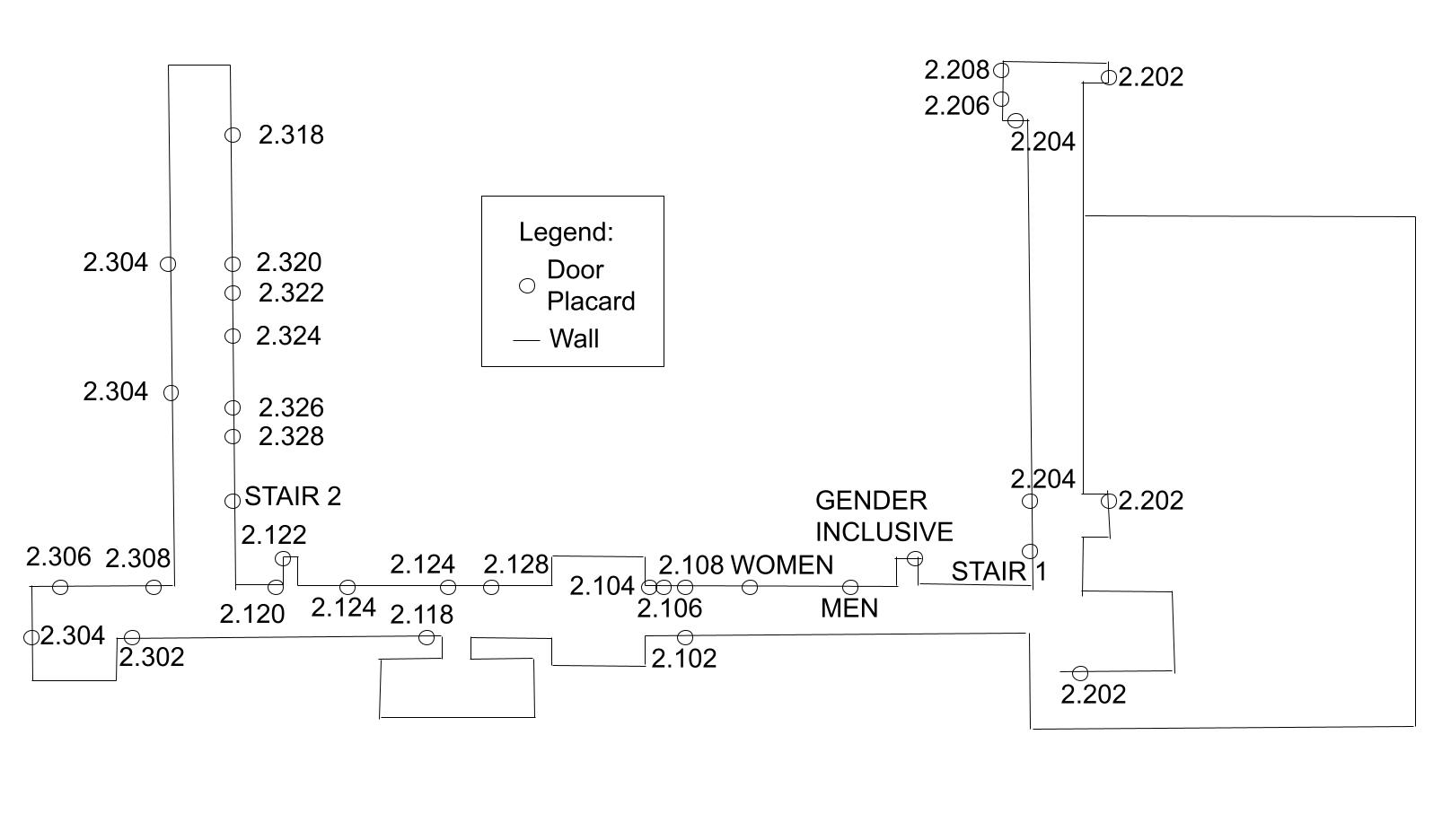}
    \caption{Hand crafted map of placard locations.}
    \label{fig:hand_crafted}
\end{figure}

Figure~\ref{fig:hand_crafted} shows a hand-crafted map of the office floor the recording was taken in. The locations of the door placards are marked by a circle, the expected labels are placed in the vicinity of the markings. Since this semantic map was hand-crafted, it is only an approximation of the land mark locations and floor plan.

The first rotation occurs at the center of the office floor and is designed to provide ORB-SLAM with an opportunity for loop-closure. This location is feature-rich and the full rotation guarantees that the robot will see a similar viewpoint when returning to the location.

The second rotation happens in the corridor on the right hand side of the map in Figure~\ref{fig:hand_crafted}, the walls on either side of the robot are nearly featureless. This second rotation is designed to test robustness to difficult localization scenarios.

The recording also roams regions of the office space which do not contain placards. The purpose is to test if the trained YOLO model provides any false positive placard detections.

\subsection{Performance Metrics}
A large portion of Sec.~\ref{sec:mappingSystem} relies on ORB-SLAM to provide reliable localization. ORB-SLAM is a stochastic process, so to properly assess the system's capabilities, multiple maps are generated by re-running ORB-SLAM on the same recording. This provides some insight into how robust the system is to unlucky mapping trials.

No record of the precise placement of the placards is available, so a baseline for numerical comparison must be established. For this, the map which subjectively appears to best represent the real world floor plan is utilized. All other maps are compared against the baseline map to assess the consistency of placard localization.

Placard poses are measured by the $(x,y,\theta)$ coordinates localized from RGB-D sensor data and ORB-SLAM's keyframe pose, where $\theta$ is the angle the vector normal to the placard's face makes with the x-axis in the 2D map projection. 

For a given observation of $\theta$, $\theta_i$, there is the true value $\theta^*_i$. Since the office floor has placards mounted on walls that lie at exclusively right angles to each other, any $\theta^*_i$ must be one of four values in increments of $\frac{\pi}{2}$. The most likely $\theta^*_i$ values are identified by finding the largest cluster of $\theta_i$ measurements in the baseline map, taking the average and then incrementing by multiples of $\frac{\pi}{2}$. Every $\theta_i$ is assigned one of these four $\theta^*_i$ values by hand labeling. 

To compare maps to the baseline poses, a correspondence between the observed placard in one map must be made to the same placard observed in the baseline. This is also done by hand labeling. If a placard is observed multiple times, but localized in different locations, every instance is marked as corresponding to the same baseline pose. Correspondences are identified through direct observation of video frames.

There are 34 placards present on the office floor, part of the evaluation will consider how many of these are missed, how many are duplicated in multiple localizations, and how many false positive placards are recorded. Of the observed true positive placards, error is computed over position and angular orientation and find as the standard deviation from the mean on each trial map.

To demonstrate the value of the enhancements to ORB-SLAM, example plots of the map with and without modification are shown with important features noted.

Finally, the portion of observed placards with correctly read labels is observed. This is determined by hand-labeling the true text on each placard.

\section{Results}
\begin{figure}
    \centering
    \includegraphics[width=\linewidth]{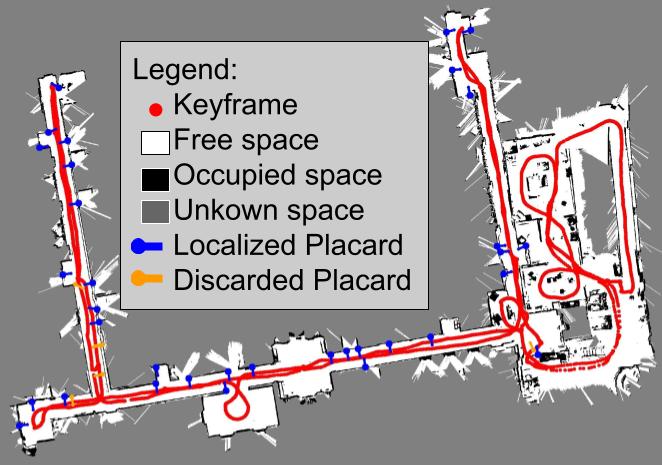}
    \caption{Trial 3 map used for baseline. (Best viewed in color.)}
    \label{fig:map3}
\end{figure}

Four ORB-SLAM maps were created to evaluate performance. Trial 3 serves as the baseline map. Figure~\ref{fig:map3} shows the map. The keyframe trajectory is shown in red. The localized placards are shown as blue dots with lines indicating their orientation. Orange markings show placards that have been discarded because they were are not localized close to a wall.

\begin{figure}
    \centering
    \includegraphics[width=\linewidth]{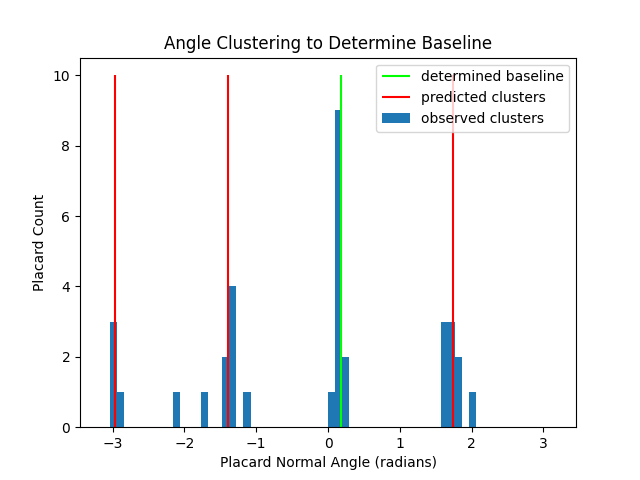}
    \caption{trial3 $\theta_i$ clustering, best viewed in color}
    \label{fig:clusters}
\end{figure}

Figure~\ref{fig:clusters} shows the clustering of $\theta_i$ values, the green vertical line indicates the selected baseline value for valid values of $\theta^*_i$, the red vertical lines indicate the predicted clusters given only $\frac{\pi}{2}$ radian angles of all the walls. As shown, the predictions align well with the observations.

%JUSTIN - DONE UP TO HERE.

Table~\ref{tab:data_summary} summarizes the numerical results. The symbols $\mu,\sigma$ here refer to the mean and standard deviation respectively of the indicated measurements. The displacement error on Trial 3 has been omitted since every placard excepting the duplicates are utilized for the baseline comparison. 

There are two placards which every trial missed, one which was the only placard not to conform to the standard measurements, and one which was posted on the window of a dark room. Since the placards are black and the dark room gave the appearance of a black wall, both of these situations presented a unique challenge to YOLOv5 which it was not trained to handle. 

Trial 4 missed one additional placard which had been discarded for not being localized close to a wall. Trial 4 and Trial 1 also falsely identify an extra placard on a section of wall with a visual similarity to the preferred objects.

Each trial contains a number of duplicate placards. Trial 2 is the worst offender because it has the worst performing map, containing segments of matching walls that are out of alignment. Placards on these walls are duplicated. Other sources of duplication come from cases where the edge of the detected placard is visually aligned with a protruding corner of a nearby wall. The resulting depth cloud includes points lying on this wall at a much closer depth thereby skewing the placard's localization.

\input{icp_merge_fig}

\begin{figure}
    \centering
    \includegraphics[width=\linewidth]{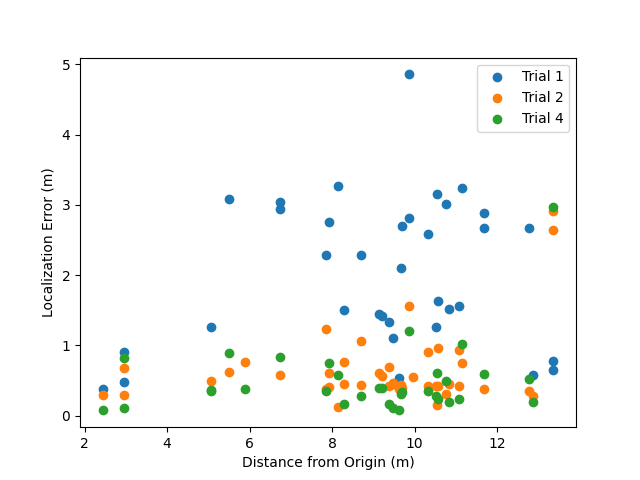}
    \caption{Scatter plot of displacement errors by distance from origin.}
    \label{fig:error_plot}
\end{figure}

Figure~\ref{fig:error_plot} shows how the displacement errors are distributed for each trial. As the placards get further away from the origin frame, defects in the map cumulatively contribute to the localization error. Therefore, the error tends to be small near the origin, and larger further away. The worst performing trial by this metric is Trial 1, while Trial 4 is the best. 

The average error in orientation remained under $6$ degrees of deviation for all trials. Again, Trial 1 has the worst results and Trial 4 the best, closely mirroring the displacement results.

\begin{table}[]
    \centering
    \begin{tabular}{|c|c|c|c|c|}
        \hline
          & trial 1 & trial 2 & trial 3 & trial 4 \\ \hline
         observed count & 35 & 43 & 34 & 34 \\ \hline
         missed count & 4 & 2 & 2 & 3 \\ \hline
         duplicates count & 5 & 11 & 2 & 2 \\ \hline
         false positive count & 1 & 0 & 0 & 1 \\ \hline
         displace error $\mu,\sigma$ (m) & 2.02,1.04 & 0.66,0.55 & - & 0.51,0.52 \\ \hline
         $\theta$ error $\mu,\sigma$ (deg) & 5.9,9.3 & 5.4,7.6 & 4.8,7.7 &  4.8,6.8\\ \hline
    \end{tabular}
    \caption{Data summary for observed placards.}
    \label{tab:data_summary}
\end{table}

Figure~\ref{fig:ICPcorrection} shows the impact of applying point cloud merging on an ORB-SLAM  map. ORB-SLAM loses tracking in the corridor on the left-hand side of Figure~\ref{fig:hand_crafted}. In this area the robot does a full rotation with blank walls on either side. Without the correction, ORB-SLAM stops tracking as it pans against each wall. Between each wall, a second map is created. After completing the rotation ORB-SLAM recognizes a return to the starting viewpoint, and begins tracking the first map again. As the robot drives back down the corridor, the localization drifts and a falsely mapped second corridor is created. When the robot returns to the half-circle viewpoint, ORB-SLAM merges the first and second maps with an added phantom hallway. However, with the modification described in Sec.~\ref{sec:modify} a single map is maintained, therefore when the robot reaches the half circle viewpoint mark ORB-SLAM instead does a loop closure which corrects the localization and merges the corridors. 

Out of the observed placards, Trial 1 correctly read the labels on $6/35$ ($17\%$), Trial 2 on $14/43$ ($33\%$), Trial 3 on $13/34$ ($38\%$), and Trial 4 on $9/33$ ($27\%$). 

\section{Conclusion}
A system for adding semantic markers to a SLAM map utilizing naturally placed text markers was presented. The system can be run on recordings of RGB-D data. The system demonstrates a high degree of consistency between mapping results as well as a robustness to featureless regions in the color video. The system is able to identify all but two text markers and read about $30\%$ of those it identifies on average.

\section*{Acknowledgment}

This work has taken place in the Learning Agents Research
Group (LARG) at the Artificial Intelligence Laboratory, The University of Texas at Austin.  LARG research is supported in part by the National Science Foundation (CPS-1739964, IIS-1724157, FAIN-2019844), the Office of Naval Research (N00014-18-2243), Army Research Office (W911NF-19-2-0333), DARPA, General Motors, Bosch, Cisco Systems, and Good Systems, a research grand challenge at the University of Texas at Austin.  The views and conclusions contained in this document are those of the authors alone.

\bibliographystyle{ieee.bst}
\bibliography{ref.bib}

\end{document}

%% file: icp_merge_fig.tex
\begin{figure*}[htb]%
    \centering
    \subfloat[\centering Without correction.\label{fig:ICPcorrection:a}]{{\includegraphics[width=.45\linewidth]{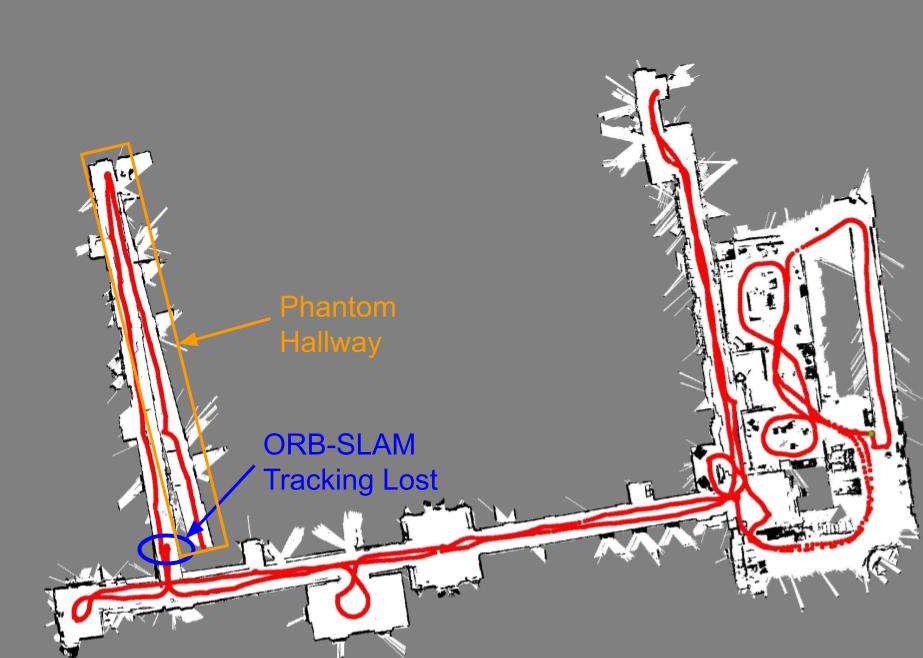} }}%
    \quad
    \subfloat[\centering With correction.\label{fig:ICPcorrection:b}]{{\includegraphics[width=.45\linewidth]{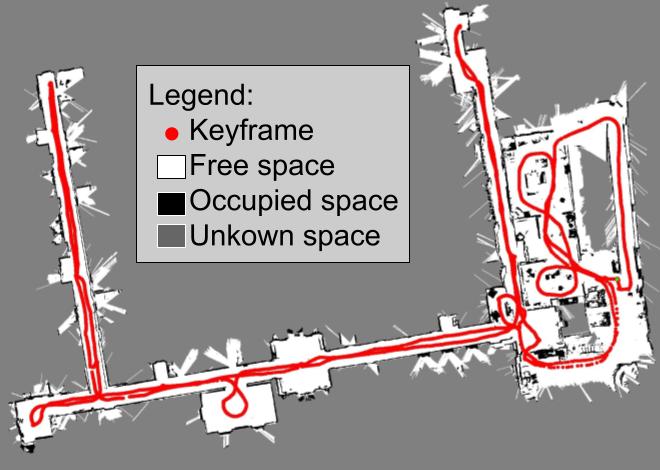} }}%
    \caption{ICP merge correction effect on ORB-SLAM trajectory.}%
    \label{fig:ICPcorrection}%
\end{figure*}